# Dataset Bias Mitigation Through Analysis of CNN Training Scores

Ekberjan Derman

Department of Computer Science, Sabancı University
Tuzla, Istanbul, Turkey
derman@sabanciuniv.edu

**Abstract**: Training datasets are crucial for convolutional neural network-based algorithms, which directly impact their overall performance. As such, using a well-structured dataset that has minimum level of bias is always desirable. In this paper, we proposed a novel, domain-independent approach, called score-based resampling (SBR), to locate the under-represented samples of the original training dataset based on the model prediction scores obtained with that training set. In our method, once trained, we use the same CNN model to infer on its own training samples, obtain prediction scores, and based on the distance between predicted and ground-truth, we identify samples that are far away from their ground-truth and augment them in the original training set. The temperature term of the Sigmoid function is decreased to better differentiate scores. For experimental evaluation, we selected one Kaggle dataset for gender classification. We first used a CNN-based classifier with relatively standard structure, trained on the training images, and evaluated on the provided validation samples of the original dataset. Then, we assessed it on a totally new test dataset consisting of light male, light female, dark male, and dark female groups. The obtained accuracies varied, revealing the existence of categorical bias against certain groups in the original dataset. Subsequently, we trained the model after resampling based on our proposed approach. We compared our method with a previously proposed variational autoencoder (VAE) based algorithm. The obtained results confirmed the validity of our proposed method regrading identifying under-represented samples among original dataset to decrease categorical bias of classifying certain groups. Although tested for gender classification, the proposed algorithm can be used for investigating dataset structure of any CNN-based tasks.

## 1 INTRODUCTION

Recently, the rapid development of deep learning (DL) greatly impacted many of the classical machine-learning based tasks, such as object detection [1], natural language processing [2], behavioral analysis [3], and robotic manipulation [4], etc. Usually, these techniques obtain performances much better than those traditional methods. The common approach to prepare such DL algorithm is to first train on a large training dataset, then evaluate its performance on a separate, previously obtained validation dataset; the final reported performance of the algorithm is based on those validation results. However, most DL-based approaches are data-hungry; that is, we need large amount of data to train them. As such, there emerged various groups and organizations to provide benchmark datasets for different DL-based tasks; they vast variety of datasets for different topics, which are usually used for benchmarking to evaluate certain algorithms. Therefore, the quality and design of these datasets are critical to reflect the actual performance of the evaluated algorithm.

In this study, we propose a new approach to investigate the structural design of the training dataset for CNN-based tasks, which is based on prediction scores of a trained model on its own training samples. The logic behind our approach can be described as follows: we believe samples with less representative groups or extreme cases obtain larger distances with their ground truth even with a trained model during training phase;

if we use such trained model to infer on its own training dataset, the distance between samples and their true labels could reflect some information regarding the quality of original training samples and their structure. Based on this information, we identify sample groups that should be upsampled and perform data augmentation on these images. The model is then trained again and evaluated on test dataset to assess the contribution of such learn-and-re-arrange approach. For evaluation, we select one gender classification dataset [6] from Kaggle [5], train a convolutional neural network (CNN) based classifier using its training images, and obtain its performance on the provided validation samples. Subsequently, we evaluate this trained classifier using a totally new test dataset consisting of four different groups, namely light male, light female, dark male, and dark female. We especially explore the prediction accuracy of the trained classifier on these groups to examine whether there exists any hidden bias of this dataset, which could be caused by its structural design that is biased against certain groups. The classification performance of the model trained using samples identified and augmented with our proposed method is compared with a variational autoencoder (VAE) based model [7] that was previously proposed for face detection for similar purpose

The remaining of this paper is organized as follows: in Section 2, we briefly introduce the related work. In Section 3, we present the approaches used for training and testing the dataset. In Section 4, the obtained experimental results are provided, followed by a discussion in Section 5. Finally, Section 6 concludes this paper.

## 2 RELATED WORK

Classification algorithms using CNN usually employ different methods to deal with data imbalance, such as pre-processing [8], in-processing to address discrimination during the model training phase [9], or post-processing, that processes the data after the model is trained [10]. These de-biasing approaches are widely used especially during the task of image classification, such as gender estimation [11].

On the other hand, as deep learning methods rely heavily on large amount of data, benchmark datasets are gaining extreme importance for the sake of reproducible research to fairly compare and evaluate different algorithms. As such, plenty of benchmark datasets were presented by various research groups and institutions around the world, such as LFW for face verification [12], FCPS for clustering [13], and MSR-VTT for video understanding [14].

A typical example could be Kaggle, an online platform to provide datasets for researchers aiming at different tasks, which has been attracting much attention especially from the deep learning community. Yu et al. proposed an SVM-based sentiment analysis on Kaggle Yelp dataset [15]. Yang et al. presented a computational framework based on deep CNN (DCNN) for Kaggle image classification dataset [16]. Pouransari et al. introduced a deep learning-based sentiment analysis on Kaggle IMDB dataset [17]. Jiang et al. proposed a credit card fraud detection algorithm based on autoencoder neural network using Kaggle credit card dataset [18]. Loey et al. introduced an automatic diagnosing method of leukemia through deep transfer learning using Kaggle blood cell image dataset [19], and the list may go on. Up to now, there exist a total of 63429 different datasets available on Kaggle, covering a vast variety of topics.

Meanwhile, bias mitigation has been attracting much attention from various research groups. Among the most recent ones, Tommasi et al. [29] presented a method to verify the potential of the DeCAF features when facing the dataset bias problem, together with a series of analyses regarding how existing datasets differ among each other. Das et al. [30] proposed a multi-task CNN to mitigate soft biometrics related bias combining age, gender and race. Li et al. [31] proposed a method to reduce dataset representational bias, called representation bias removal (REPAIR) procedure. Their method seeks a weight distribution that penalizes examples easy for a classifier built on a given feature representation, and maximizes the ratio between the classification loss on the reweighted dataset and the uncertainty of the ground-truth labels. Wang et al. [32] designed a simple yet effective visual recognition benchmark to compare bias mitigation techniques.

At the same time, debiasing has also attracted some attention from the Kaggle community. Alexandru Papiu published a Kaggle post regarding bias correction with XGBoost [20]. Rachael Tatman opened a discussion on Kaggle regarding gender bias in word embeddings [21]. In 2019, Kaggle launched a competition named 'Jigsaw Unintended Bias in Toxicity Classification', which has attracted a total of 3165 teams' participation [22]. Early in this year, there is another competition launched on Kaggle regarding the gender bias detection in social media comments in South Korea [23], which has attracted 5 teams up to now. Obviously, datasets and works on bias detection have been continuously gaining popularity among the research community.

## 3 METHODOLOGY

In this study, our main objective is to present a domain-independent approach to identify under-represented samples in the training dataset for CNN-based classification tasks. To evaluate performance, we targeted a gender classification as an example using a Kaggle dataset. As such, we will not focus on performance improvement of the classification algorithm itself, but rather use a typical CNN structure to perform our classification; subsequently, we use a ready-to-use variational autoencoder proposed by [7] to include the underlying latent space of the training data to explore its effect on classification performance. Moreover, we propose a totally new approach, named score-based resampling (SBR), to identify which samples should be upsampled in the original training dataset based on prediction of a trained network, augment them and train our classifier again; for final classification, we introduce RBF kernel-based SVM on top of our CNN output.

### 3.1 CNN Based Classifier

Based on [7] and its implementation [24], our CNN-based classifier has a relatively standard architecture, consisting of a series of convolutional layers with batch normalization, followed by two fully connected layers to flatten the convolution output and generate a class prediction, which is illustrated in Figure 1. We used a batch size of 32, total epoch of 20, number of filters as 12, and Adam optimizer with a learning rate of 0.0001, alpha as 0.9, beta as 0.999, and epsilon as $10^{-8}$. Each image is normalized and resized to 64x64 before passing to the CNN. One main difference between our CNN structure and that in [24] is that, we removed batch normalization layers, as the CNN has relatively simple structure, and smaller training epochs.

### 3.2 Algorithmic Bias Mitigation

Imbalance in training data can lead to unexpected algorithmic bias. For instance, if the training data for gender classification consists mainly of light-skinned males, or the proportion of males is significantly larger than that of females, training using such dataset could result in a classifier that could be better suited at recognizing and classifying facial images with similar features, and thus, will be biased. To overcome this difficulty, a naïve solution would be to annotate different subclasses (such as light-skinned females, dark-skinned males, etc.) within the training data, and then manually even out the data with respect to these groups. However, this approach suffers from two main disadvantages: first, it requires massive amount of data to be annotated; second, we need to know the potential biases a priori. Thus, such manual annotation could fail to capture all the different features that are imbalanced among the training data.

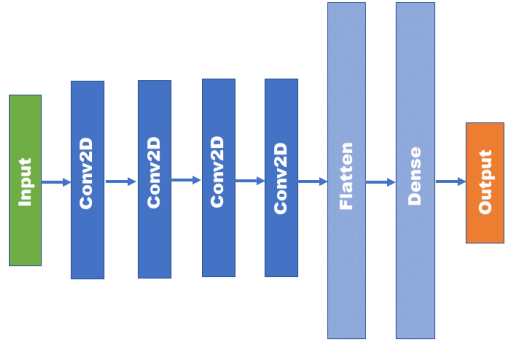

Figure 1. The model structure of our CNN classifier. This structure is a relatively standard one, and was used by authors of [7] in their implementation [24]. For consistency, we also adopted the same model in this work.

Therefore, in this work, by comparing an existing variational autoencoder against our newly proposed distance-based approach, we decide to learn these features in an unbiased, unsupervised manner, thus, removing the need for any annotation, followed by training a classifier fairly with respect to such features.

### 3.3 Debiasing Variational Autoencoder (DB-VAE)

Amini et al. [7] proposed a variational autoencoder based debiasing approach to investigate potential hidden bias for face detection regarding different groups of skin colors. In their approach, a model was trained to learn a representation of the underlying latent space of the training data, and use this information to mitigate unexpected biases during training, by sampling images from under-represented groups more frequently.

To learn the latent space, we can constrain the means and standard deviations of our VAE to approximately be a unit Gaussian. The loss function includes two terms: a latent loss ($L_{KL}$) that measures how closely the learned latent variables match a unit Gaussian, which is defined by the Kullback-Leibler (KL) divergence; and a reconstruction loss ($L_x(x, \hat{x})$) that measures how accurately the reconstructed outputs match the input. These loss functions can be described in the following equations:

$$L_{KL}(\mu, \sigma) = \frac{1}{2}\sum_{j=0}^{k-1}(\sigma_j + \mu_j^2 - 1 - log\sigma_j) \quad (1)$$

$$L_x(x, \hat{x}) = ||x - \hat{x}||_1 \quad (2)$$

Therefore, the VAE loss can be expressed as:

$$L_{VAE} = c \cdot L_{KL} + L_x(x, \hat{x}) \quad (3)$$

in which, $c$ represents a regularization coefficient.

The VAE encoder generates vector of means and standard deviations respectively, that are constrained to roughly follow Gaussian distributions. A sampled latent variable $z$ can be formalized as follows:

$$z = \mu + e^{(\frac{1}{2} \cdot log\Sigma)} \cdot \epsilon \quad (4)$$

where $\mu$ is the mean and $\Sigma$ represents the covariance matrix.

Based on this VAE architecture, a debiasing VAE can be constructed as shown in Figure 2.

The encoder in the DB-VAE outputs a single supervised variable, $z_o$, that corresponds to the class prediction – male or female. The loss function of DB-VAE can be expressed as follows:

$$L_{total} = L_y(y, \hat{y}) + I_f(y)[L_{VAE}] \quad (5)$$

in which, $I_f(y) = 0$ and $I_f(y) = 1$ correspond to indicator variables of male and female images.

As the input images are given through the network, the encoder learns an estimate $\mathcal{Q}(z|x)$ of the latent space. The relative frequency of rare data is increased through increased sampling of under-represented regions of the latent space. As such, $\mathcal{Q}(z|x)$ can be approximated with the frequency distributions of each of the learned latent variables, followed by the definition of the probability distribution of selecting a given datapoint $x$ based on this approximation. These obtained probability distributions are used during training to re-sample the data

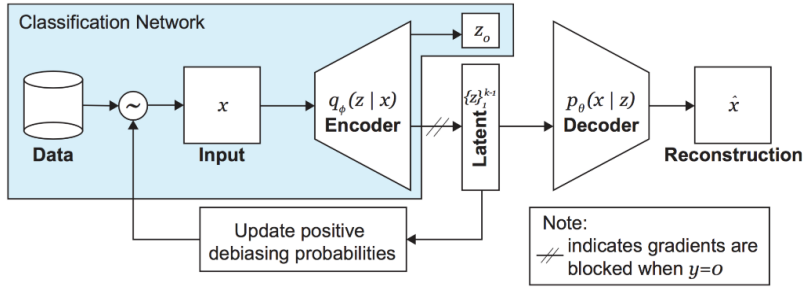

Figure 2. The architecture of DB-VAE (image adopted from [7])

### 3.4 Score-Based Resampling (SBR)

#### 3.4.1 Investigating Training Space

The gradient descent-based weight-update mechanism of a CNN model is depending on the chosen loss function. Usually, the loss function reflects the distance between the predicted and true labels of a sample. As such, once a CNN model is trained and its weights are finalized, if we investigate the scores of the last layer on the training data, we can actually build a vision regarding the internal structure of the training dataset; in the final weight-update training stage, a CNN model produces much higher score for samples it learnt well, while generates relatively lower scores for those challenging or under-represented samples from the training dataset.

Instead of the final classification label of the model on its training data after weight finalization, our main focus is the distance or score of each training samples. For instance, after training, the CNN model can generate scores of 0.0001 and 0.25 for two images with true labels both as '0'. If we focus on class labels only, these can all be classified correctly with a good threshold value, such as 0.5; however, this is not informative for our objective. But instead, the two scores reveal that the first image is among the majority of samples that our model learnt well, while it is highly possible that the second one is from under-represented samples. As such, we can upsample such images to reconstruct our training dataset, and train our model again on it. In this manner, our model can learn both the originally well-represented, as well as under-represented images.

In this work, we used a distance threshold of 0.15 to determine those samples that need to be upsampled; that is, any sample having distance higher than 0.15 with its true label will be marked as under-represented, and will be upsampled using data augmentation method. The steps of our SBR can be briefly summarized as in Algorithm 1.

#### 3.4.2 Data augmentation

Once we identify those under-represented images as mentioned above, we can perform data augmentation on them. Here in this work, we perform image flipping, and cropping the 85% of the image starting from left, as well as right, thus, leading to three more different versions of an original under-represented sample; that is, we upsample those less representative images by three times more to construct our new training dataset.

#### 3.4.3 Temperature in Sigmoid Function

In neural networks, class probabilities are typically produced by a "softmax" layer that converts the logit, $z_i$, into a probability, $q_i$, for each class, by comparing $z_i$ with other logits [25]:

$$q_i = \frac{\exp(\frac{z_i}{T})}{\sum_j \exp(\frac{z_j}{T})} \quad (6)$$

in which, *T* is a temperature that is normally set to 1. Higher *T* value generates a softer probability distribution over classes.

Similarly, in our CNN model, for sigmoid output, we can introduce such temperature term to make more softer or shaper score transition as well:

$$\sigma(x) = \frac{1}{1+\exp\left(-\frac{x}{T}\right)} \qquad (7)$$

In this work, we selected T = 0.85 to make the score distributions of two classes become more easily separable so that we can better identify samples receiving lower scores.

### 3.4.4 CNN+SVM Classification

In our proposed SBR, for final classification, instead of determining labels based on predefined hard thresholds, such as 0.5 as that in [24], we incorporate an RBF-based SVM classifier on top of our CNN layer. That is, once we get the scores of final sigmoid layer, we train our SVM classifier based on them; during prediction, the input to our SVM classifier is the sigmoid output of our CNN model. We used five-fold cross validation technique to determine the best gamma and C values, which were selected as 0.5 and 10.0, respectively.

| **Algorithm 1: SBR Steps** |
| --- |
| 1. Train CNN model using the original training images.
2. Once trained, use the model to perform inference on the same training dataset.
3. Obtain the score of sigmoid layer of CNN model for each training sample.
4. Threshold the obtained scores to determine those images with lower scores than others within the same class, mark them as our target (under-represented samples).
5. Augment those under-represented images by flipping and crop operations, then add to our original training dataset.
6. Train a new CNN model on this newly organized training dataset.
7. Train an RBF kernel-based SVM classifier using scores of the newly trained CNN model, then incorporate it on top of our CNN model to perform final class prediction, parameters of RBF kernel are determined by five-fold cross validation. |

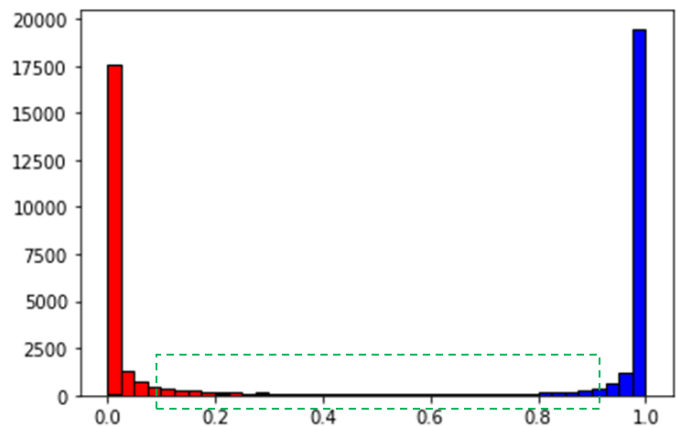

Figure 3. Score distribution of our trained CNN model on original training data. Red: samples with true label as 0. Blue: samples with true label as 1. Samples with scores in the region marked with dashed rectangle are of our main interest.

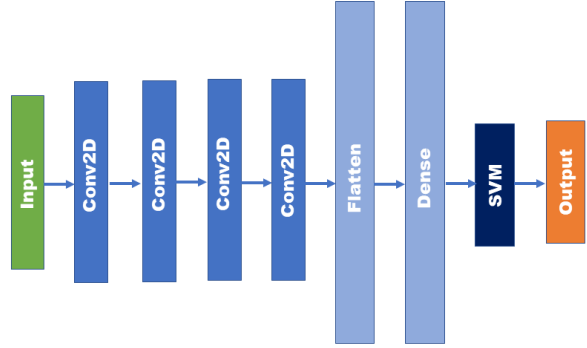

Figure 4. Illustration of our CNN model with SVM classifier.

### 3.4.5 Facial Detection and Extraction

As it can be observed from figures 5 and 6, we noticed that our test images are different from our training images in the form of facial region; that is, our training data is of face images, while our constructed test images include individual's face as well as other background information. As such, we also decided to first perform facial region detection and extraction on our test images, then perform corresponding CNN inference on them. In this manner, we believe those negative effects introduced by regions other than face could be decreased, leading to more accurate evaluation of our investigated approaches. Here, we used Dlib [28] to detect facial regions.

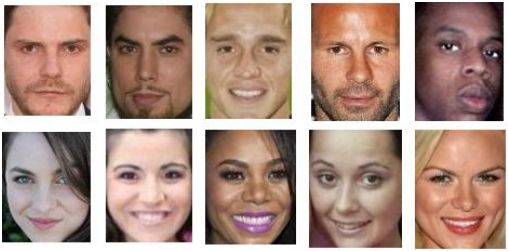

Figure 5. Sample images of our selected training dataset. Upper: males. Lower: females.

## 4 EXPERIMENTAL EVALUATIONS

### 4.1 Dataset

We trained our CNN classifier using the training images of the gender classification dataset from Kaggle [6], and evaluated it on the validation images of the same dataset. This dataset consists of 23,766 and 23,243 training images, as well as 5808 and 5841 validation images for male and female classes, respectively.

Moreover, we constructed a totally new test dataset using the CelebAMask-HQ dataset [12], which contains large-scale face images with 30,000 high-resolution face images. Our constructed dataset consists of four different groups, namely light male (lm), light female (lf), dark male (dm), and dark female (df), each group with a total of 705, 709, 604, and 718 images, respectively.

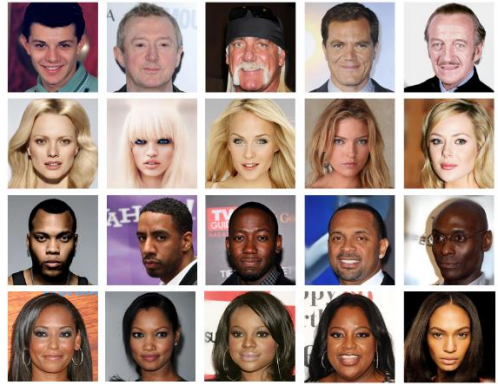

Figure 6. Sample images from our constructed test dataset. From top to bottom: light males, light females, dark males, and dark females.

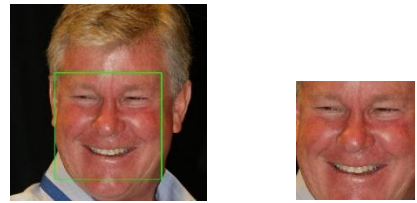

Figure 7. Sample image from our test dataset with detected facial region marked as green rectangle (left), and its extracted facial image (right)

## 4.2 Implementation

Our implementation is based on Python language, using Tensorflow framework [26] in Google Colab environment [27]. Our CNN model uses a total number of filters as 12, initial learning rate as 0.0001, as well as Adam optimizer with first and second moment estimates (beta_1 and beta_2) as 0.9 and 0.999, respectively., while the DB-VAE is directly adopted from its original implementation [24]. As such, we used a total of 100 latent variables, with 20 epochs, latent regularization parameter as 0.0005, and the same Adam optimizer that is used in our CNN classifier training.

For determining best parameters for our RBF kernel-based SVM, we evaluated values of gamma and C as $[2^{-4}, 2^{-3}, 2^{-2}, 2^{-1}, 2^{0}]$ and $[10^{-4}, 10^{-3}, 10^{-2}, 10^{-1}, 10^{0}, 10^{1}, 10^{2}]$, respectively. In both cases, the total epochs of our CNN models were set to 20 in order to keep the same with that of DB-VAE. Meanwhile, we also applied early stopping criteria for both of our CNN training phases: the model would stop training if the training accuracy fails to improve within five consecutive epochs.

## 4.2 Results

There were totally 3437 images identified as from under-represented groups using our SBR approach, and selected to be augmented, leading to increasing the size of training dataset from 47009 to 57320. Some of those samples are demonstrated in Figure 7.

The accuracies of original standard CNN, DB-VAE, and CNN+SVM trained on the newly revised training dataset using our SBR approach on the original validation, as well as our test datasets, are presented in Table 1 and Figures 9-11, respectively.

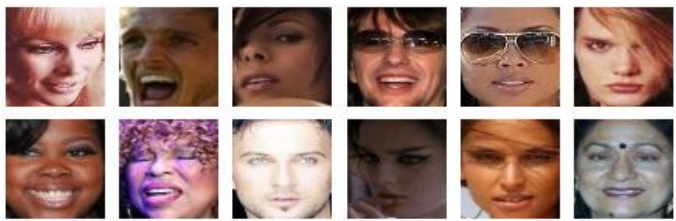

Figure 8. Sample images that were identified by our SBR as needed to be augmented.

Table 1. Obtained Accuracies of Standard CNN, DB-VAE, and CNN+SVM methods

| Datasets/Methods | | Standard CNN | DB-VAE | CNN+SVM (our approach) |
|---|---|---|---|---|
| Validation Set | | 0.9523 | 0.9585 | 0.9482 |
| Test Set | original | 0.6389 | 0.4919 | 0.5789 |
| | with face extraction | 0.7235 | 0.4871 | **0.7546** |
| LM | original | 0.7560 | 0.7007 | 0.3617 |
| | with face extraction | 0.9901 | 0.9844 | 0.6326 |
| LF | original | 0.3512 | 0.5007 | 0.7080 |
| | with face extraction | 0.5078 | 0.4866 | **0.8956** |
| DM | original | 0.8758 | 0.8940 | 0.5364 |
| | with face extraction | 0.9668 | 0.9867 | 0.6362 |
| DF | original | 0.6086 | 0.3106 | 0.7006 |
| | with face extraction | 0.4708 | 0.2577 | **0.8370** |

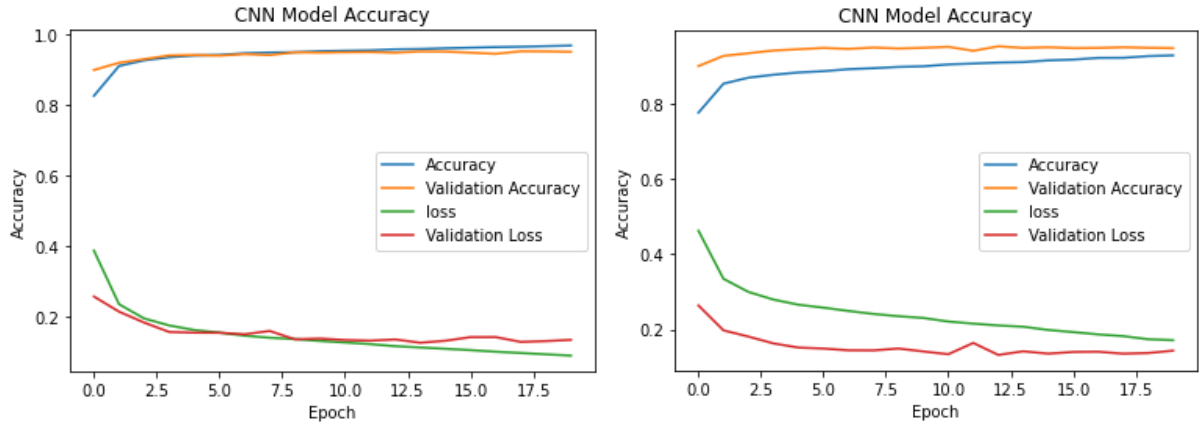

Figure 9. CNN Model accuracy and loss on training and validation dataset. Left: training with original training dataset. Right: training with our SBR-based newly augmented training dataset.

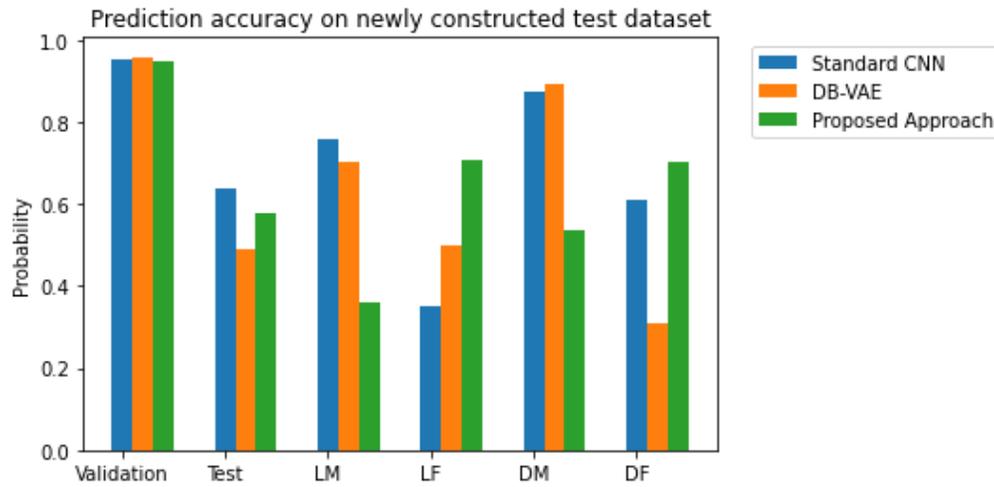

Figure 10. Prediction accuracies of standard CNN classifier, DB-VAE, and our CNN+SVM classifier on our newly constructed test dataset (without face detection and extraction step).

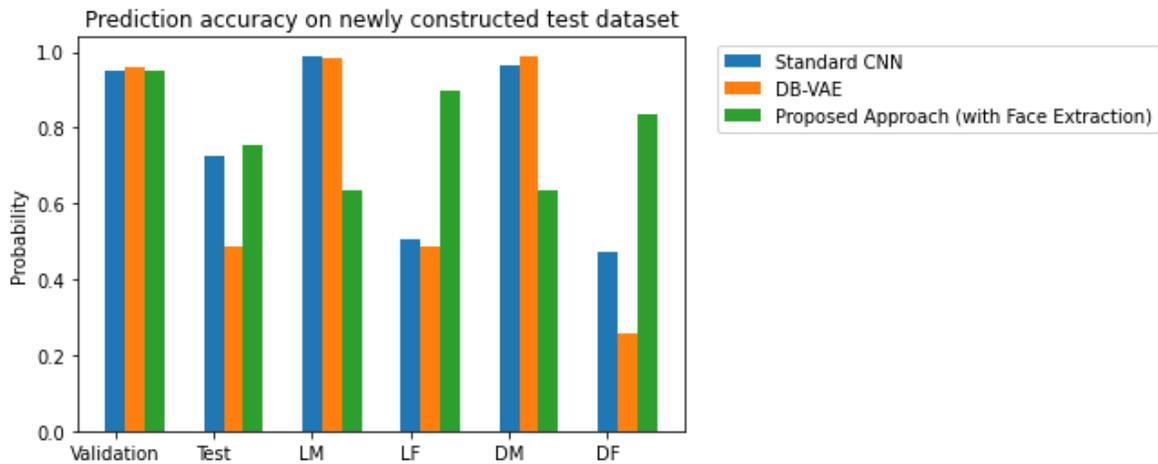

Figure 11. Prediction accuracies of standard CNN classifier, DB-VAE, and our CNN+SVM classifier on our newly constructed test dataset (with face detection and extraction step).

## 5 DISCUSSION

As it can be seen from the obtained results, the trained standard CNN performs well on the validation images of our investigated dataset, but demonstrates inferior performance on our newly constructed test dataset. This demonstrates that the investigated gender classification dataset is not representative, as such, algorithms tested on this dataset and reported their corresponding performances are not generalizable and need to be further examine.

Moreover, while all the three approaches achieved similar performances on the original validation dataset, DB-VAE performed inferior to the standard CNN and our proposed approach on the test dataset. After facial region extraction, we have successfully removed unwanted background information from our test images; as such, our proposed approach achieved the best accuracy. Regarding specific groups, DB-VAE achieved performance improvement on dark male groups; on the other hand, our proposed approach has successfully improved the classification performance of light and dark females while decreasing accuracy for light and dark males, indicating that those upsampled images are more likely to be from female groups. Also, there exist large margins regarding performance differences of our approach with that standard CNN and DB-VAE. Meanwhile, while obtaining performance improvement on overall test data, our approach has demonstrated significant performance decrease regarding male groups; this indicates the newly augmented test data has successfully increased the sample size of under-represented female groups more than male groups, which also gives us information regarding the structural design of our investigated dataset, that includes more female images of extreme cases than male images.

Combining Figures 8, 10 and 11, as well as Table 1, we can observe that the main reason for making the investigated dataset to be less representative is not due to the skin color differences, but rather caused by other factors such as occlusion (sunglasses or hair), lighting conditions, head pose, and facial expression, etc. Therefore, even though it was utilized in [7] well for face detection and inspired us to carry out this work, evaluating merely based on skin colors to determine the under-represented samples is not appropriate for this case. Instead, increasing samples with occlusion, different illumination conditions, as well as pose and facial expressions in the original training dataset could positively contribute to improve the generalizability of the investigated dataset.

On the other hand, we can state that our proposed SBR approach is valid to identify those less representative samples in the original dataset. We believe that this approach could also be introduced to other tasks than gender classification or face detection. Especially, utilizing the scores of trained models on their own training samples to investigate the structure of the training dataset can be effectively used for any other CNN-based classification task. Meanwhile, combining SVM with CNN to replace any pre-defined thresholding is applicable to decide the classification margin more effectively than hard-thresholding.

## 6 CONCLUSION AND FUTURE WORK

In this study, we proposed a domain-independent method to identify under-represented samples among training dataset for CNN-based tasks by using the model prediction score; once a CNN model was trained, inference was performed on the original training samples, and based on the distance between true and predicted labels, we determined samples that needed to be upsampled, and augmented the training dataset; subsequently, the CNN model was trained from scratch on this new dataset. For experimental evaluation, we chose a Kaggle gender classification dataset, and carried out performance evaluation between one CNN model with general structure, one based on [7], and one trained with revised datasets based on our proposed method. During our experiments, we incorporated RBF kernel-based SVM with our CNN model for prediction. Experimental results demonstrated the effectiveness of our proposed method, as well as the fact that the investigated dataset cannot be used for general purpose, as there was hidden bias due to its structural design.

Although we evaluated our proposed approach on gender classification, it can be used for any CNN-based tasks to investigate the structural design of the training dataset, investigate hidden bias, and identify those less representative samples to further processing, such as upsampling.

As our future work, we could investigate the possibility of employing our proposed SBR approach to identify under-represented samples in other domains using CNNs, such as classification or detection of various objects, financial market analysis, or anomaly detection, etc. Meanwhile, incorporating distribution analysis using the identified under-represented samples could lead to more informative results regarding the internal structure of the target dataset.

**References**

[1] Z.Q. Zhao, P. Zheng, S.T. Xu, and X. Wu, "Object Detection with Deep Learning: A Review", *Computer Vision and Pattern Recognition*, *arXiv:1807.05511,* 16 April, 2019.

[2] T. Young, D. Hazarika, S. Poria, and E. Cambria, "Recent Trends in Deep Learning Based Natural Language Processing", *Computation and Language*, *arXiv:1708.02709*, 25 November, 2018.

[3] O. Sturman, L.V. Ziegler, C. Schlappi, F. Akyol, M. Privitera, D. Slominski, C. Grimm, L. Thieren, V. Zerbi, B. Grewe, and J. Bohacek, "Deep Learning Based Behavioral Analysis Reaches Human Accuracy and Is Capable of Outperforming Commercial Solutions", *Neuropsychopharmacology*, Nature, 25 July, 2020.

[4] O. Kroemer, S. Niekum, and G. Konidaris, "A Review of Robot Learning for Manipulation: Challenges, Representations, and Algorithms", *Robotics, arXiv:1907.03146*, 6 November, 2020.

[5] www.kaggle.com

[6] Kaggle gender classification dataset, available at: https://www.kaggle.com/cashutosh/gender-classification-dataset

[7] A. Amini, A. P. Soleimany, W. Schwarting, S. N. Bhatia, and D. Rus, "Uncovering and Mitigating Algorithmic Bias Through Learned Latent Structure", *AIES'19: Proceedings of the 2017 AAAI/ACM Conference on AI, Ethics, and Society*, pp:289-295, January 2019.

[8] L. E. Celis, V. Keswani, and N. K. Vishnoi, "Data Preprocessing to Mitigate Bias: A Maximum Entropy Based Approach", *Machine Learning, arXiv:1906.02164*, 30 June, 2020.

[9] B. d'Alessandro, C. O'Neil, and T. LaGatta, "Conscientious Classification: A Data Scientist's Guide to Discrimination-Aware Classification", *Machine Learning, arxiv:1907.09013*, 21 July, 2019.

[10] P. Awasthi, M. Kleindessner, and J. Morgenstern, "Equalized Odds Postprocessing Under Imperfect Group Information", *Machine Learning, arXiv: 1906.03284*, 2 March, 2020.

[11] A. H. Kamarulzalis, M. H. M. Razali, and B. Moktar, "Data Pre-processing Using SMOTE Technique for Gender Classification with Imbalance Hu's Moments Features", *In Proceedings of the Second International Conference on the Future of ASEAN (ICoFA)*, vol:2, pp: 373-379, 2017.

[12] C.H. Lee, Z. Liu, L. Wu, and P. Luo, "MaskGAN: Towards Diverse and Interactive Facial Image Manipulation", *IEEE Conference on Computer Vision and Pattern Recognition (CVPR)*, 2020.

[13] A. Ultsch, "Clustering with SOM: U*C", *In Proceedings of Workshop on Self-organizing Maps*, pp: 75-82, Paris, France, 2005.

[14] J. Xu, T. Mei, T. Yao, and Y. Rui, "MSR-VTT: A Large Video Description Dataset for Bridging Video and Language", *IEEE Conference on Computer Vision and Pattern Recognition (CVPR)*, 27-30 June, 2016.

[15] B. Yu, J. Zhou, Y. Zhang, and Y. Cao, "Identifying restaurant features via sentiment analysis on Yelp Reviews", *Computation and Language*, arXiv: 1709.08698, 20 September, 2017.

[16] X. Yang, Z. Zeng, S. Teo, L. Wang, V. Chandrasekhar, and S. Hoi, "Deep Learning for Practical Image Recognition: Case Study on Kaggle Competitions", *24th ACM SIGKDD Conference on Knowledge Discovery and Data Mining*, London, UK, 19-23 August, 2018.

[17] H. Pouransari and S. Ghili, "Deep Learning for Sentiment Analysis of Movie Reviews", *Technical Report*, *Stanford University*, 2014.

[18] J. Zou, J. Zhang, and P. Jiang, "Credit Card Fraud Detection Using Autoencoder Neural Network", *Machine Learning, arXiv:1908.11553*, 30 August, 2019.

[19] M. Loey, M. Naman, and H. Zayed, "Deep Transfer Learning in Diagnosing Leukemia in Blood Cells", *MDPI Journal of Computers, vol:9, issue:2*, 15 April, 2020.
[20] https://www.kaggle.com/apapiu/bias-correction-xgboost
[21] https://www.kaggle.com/rtatman/gender-bias-in-word-embeddings
[22] https://www.kaggle.com/c/jigsaw-unintended-bias-in-toxicity-classification/overview
[23] https://www.kaggle.com/c/korean-bias-detection
[24] https://github.com/aamini/introtodeeplearning/blob/master/lab2/solutions/Part2_Debiasing_Solution.ipynb
[25] G. Hinton, O. Vinyals, and J. Dean, "Distilling the Knowledge in a Neural Network*", NIPS 2014 Deep Learning Workshop*.
[26] Abadi, Martin et al., 2016. Tensorflow: A system for large-scale machine learning. *In 12th Symposium on Operating Systems Design and Implementation*. pp. 265–283.
[27] https://colab.research.google.com
[28] D. E. King, "Dlib-ml: A Machine Learning Toolkit", *Journal of Machine Learning Research*, vol: 10, pp: 1755-1758, 2009.
[29] T. Tommasi, N. Patricia, B. Caputo, and T. Tuytelaars, "A Deeper Look at Dataset Bias", *arXiv:1505.01257v1*
[30] A. Das, A. Dantcheva, and F. Bremond, "Mitigating Bias in Gender, Age, and Ethnicity Classification: a Multi-Task Convolution Neural Network Approach", *ECCVW*, Sep 2018, Munich, Germany. hal-01892103
[31] Y. Li, and N. Vasconcelos, "Dataset Resampling", *Proceedings of the IEEE/CVF Conference on Computer Vision and Pattern Recognition (CVPR)*, 2019, pp. 9572-9581
[32] Z. Wang, K. Qinami, I. C. Karakozis, K. Genova, P. Nair, K. Hata, and H. Russakovsky, "Towards Fairness in Visual Recognition: Effective Strategies for Bias Mitigation", *CVPR2020*, arXiv:1911.11834v2